\documentclass[10pt,twocolumn,letterpaper]{article}

\usepackage{cvpr}              

\usepackage{graphicx}
\usepackage{amsmath}
\usepackage{amssymb}
\usepackage{booktabs}
\usepackage{multirow}
\usepackage{arydshln}

\usepackage[pagebackref,breaklinks,colorlinks]{hyperref}

\usepackage[capitalize]{cleveref}
\crefname{section}{Sec.}{Secs.}
\Crefname{section}{Section}{Sections}
\Crefname{table}{Table}{Tables}
\crefname{table}{Tab.}{Tabs.}

\def\cvprPaperID{*****} 
\def\confName{CVPR}
\def\confYear{2022}

\begin{document}

\title{PixTrack: Precise 6DoF Object Pose Tracking using NeRF Templates and Feature-metric Alignment}

\author{Prajwal Chidananda \quad \quad Saurabh Nair \quad \quad Douglas Lee \quad \quad Adrian Kaehler \\ 
Giant AI, Inc.\\
}

\makeatletter
\let\@oldmaketitle\@maketitle
\renewcommand{\@maketitle}{\@oldmaketitle

  \includegraphics[width=\linewidth]
    {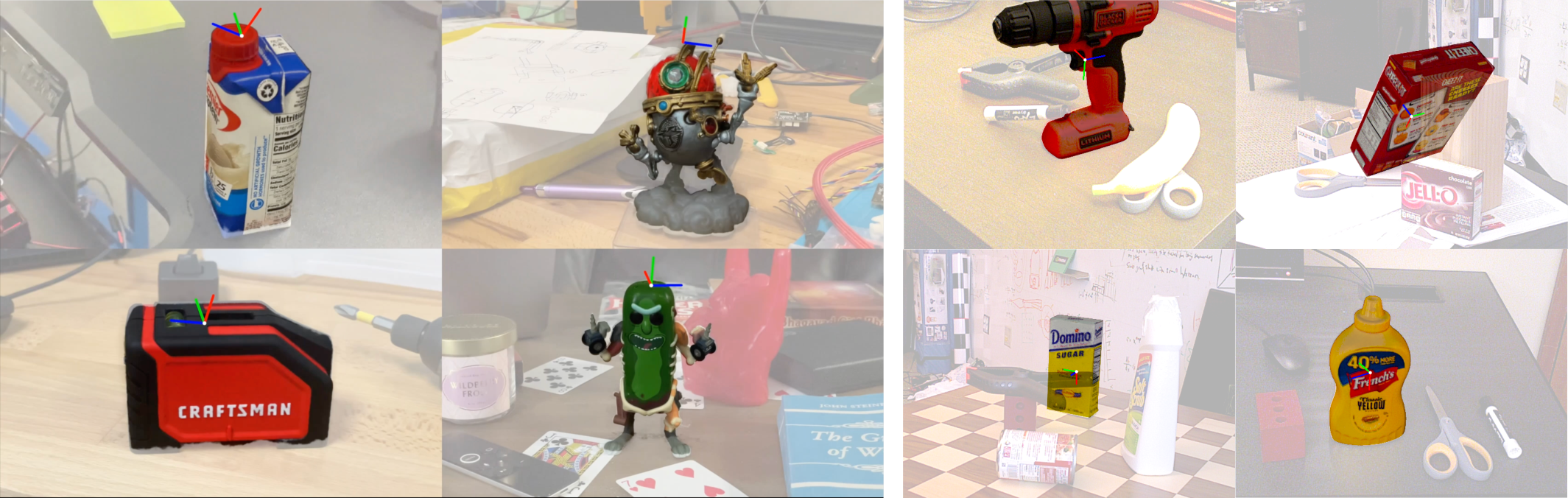}
    \captionof{figure}{PixTrack uses a Neural Radiance Field as the canonical representation of a given object and provides pixel-level accuracy in 6-DoF object tracking for monocular RGB (images on the left) or RGB-D (images on the right) sequences.}
    
    \bigskip}

\makeatother

\maketitle

\begin{abstract}
We present PixTrack, a vision based object pose tracking framework using novel view synthesis and deep feature-metric alignment. We follow an SfM-based relocalization paradigm where we use a Neural Radiance Field to canonically represent the tracked object. Our evaluations demonstrate that our method produces highly accurate, robust, and jitter-free 6DoF pose estimates of objects in both monocular RGB images and RGB-D images without the need of any data annotation or trajectory smoothing. Our method is also computationally  efficient making it easy to have multi-object tracking with no alteration to our algorithm through simple CPU multiprocessing. Our code is available at: \href{https://github.com/GiantAI/pixtrack}{https://github.com/GiantAI/pixtrack} 
\end{abstract}

\section{Background}
Accurately detecting and tracking the SE(3) poses of objects in a scene, is crucial for applications like grasping, object manipulation in robotics, augmented reality, and path planning in autonomous mobility. This is a hard problem given the variance in lighting, backgrounds, symmetry and the presence of other objects. 

Most classical methods rely on creating a canonical 3D keypoint model of the object using a method like structure from motion, and then running PnP+Ransac on query RGB frames. These methods rely on the keypoint estimation to be accurate, which are susceptible to large lighting changes, distractors etc. 
RGB based object pose estimation has been, to a substantial degree, addressed by Deep Learning based methods such \cite{kaiminghe_MaskRCNN} and DeepLabv3\cite{deeplabv3plus2018}. Estimating 6DoF pose represents a greater challenge as the canonical model of the object needs to be encoded in estimation pipeline in a certain way. Usually the 2D detectors predict keypoints that get lifted to 3D using PnP; We believe this approach is ill-posed, as the detector doesn't have any prior understanding of 3D geometry. 

\label{sec:intro}

\section{Related Work}
\label{subsec:related_work}

\begin{figure*}[h!]
\centering
\includegraphics[width=\textwidth]{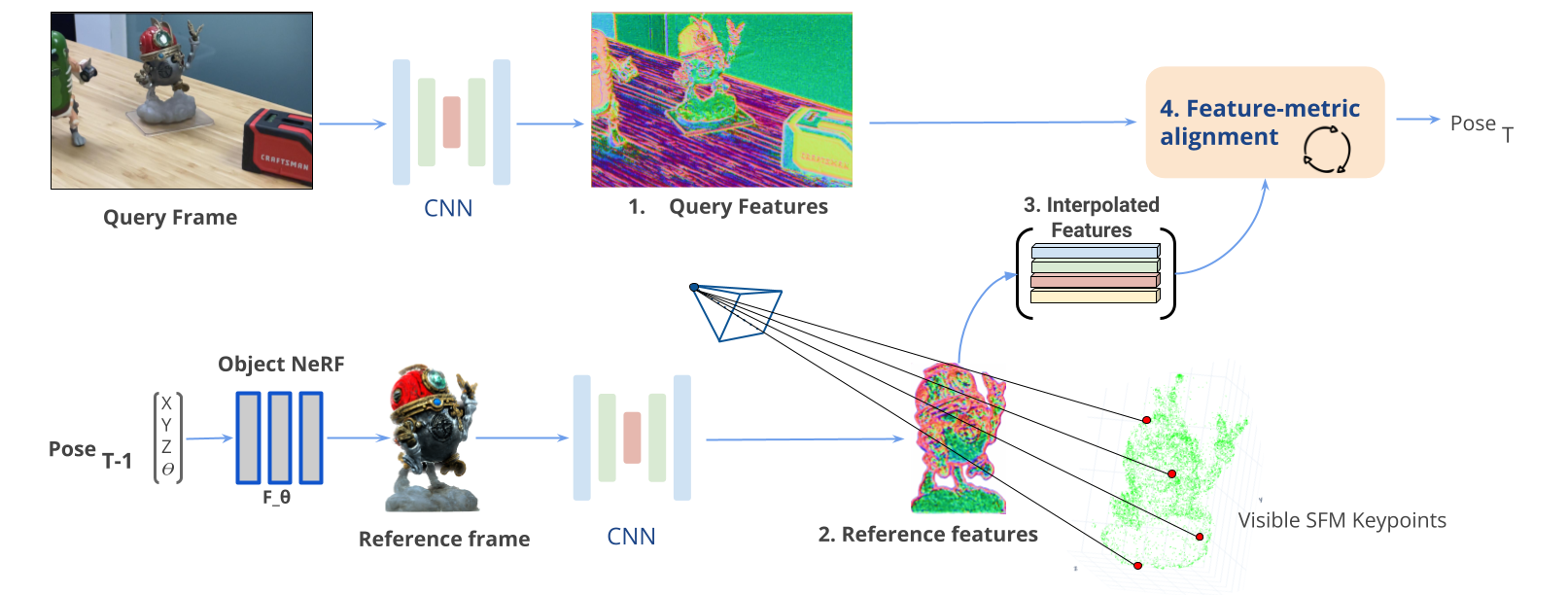}
\caption{Localization technique used in PixTrack. 1. Input frame feature extraction: Given an input RGB frame, we extract deep features using a pre-trained CNN. 2. Reference frame feature extraction: Using the predicted pose from the previous frame, we render a novel view using a canonical object-NeRF. We use this view as the reference frame and extract deep features using the aforementioned pre-trained CNN. 3. Reference frame feature interpolation: We interpolate the reference frame features at 2D points that were obtained by projecting the 3D points from the SfM onto the reference image. 4. Feature-metric alignment: We iteratively optimize using an LM optimizer over SE(3) to obtain the pose of the object in the input frame}
\end{figure*}
\label{sec:pixtrack}

RNNPose\cite{RNNPose} uses feature optimization and novel view rendering. Their method uses a CAD model of the object, from which the novel view is derived. NeRF rendered images are proven to be more representative of the object's projection than a CAD model that has limited resolution. 

Pose estimation using NeRFs\cite{mildenhall2020nerf} was first demonstrated by iNeRF\cite{iNeRF}, which estimates the pose of a new view using gradient based optimization on image pixels rendered from the NeRF at a specified camera pose. This method is non-feature based, limited to relatively small viewpoint changes and thus requiring a good initial estimate of the pose, and suffers from high inference latency which precludes online operation. Moreover, iNeRF only demonstrated operations on synthetic 360° (NeRF) and small baseline front facing real (LLFF) datasets.

 NeRF-Pose \cite{NeRFPose} uses NeRF templates, a pose regression DNN, and a RANSAC + PnP pose refinement step. This architecture suffers from limitations due to domain discrepancies between the weak supervision data used to train the pose-network and the NeRF generated images leading to suboptimal solutions which RANSAC+PnP struggles to refine. Our (PnP-free) method does not require training an auxiliary network to perform pose inference. Our evaluations also suggest the deep-feature metric alignment is much more robust than Ransac+PNP over 3D keypoints.

OnePose\cite{OnePose} a one shot pose estimation method from the query images directly is feature based, but does not exploit novel view synthesis. It also adds the necessity to maintain a database of reference frames and their features. This method also requires a human annotation of 2D keyframes, which our method doesn't require. 

Our approach is very close to \cite{wen2023bundlesdf}, which uses neural-rendering to optimize over 3D points SFM and depth, but our method adds a feature-metric alignment step that makes Pixtrack work optionally with just RGB images.

\section{Motivation: Tracking as SFM-localization}
We treat object pose tracking as an SfM-localization problem. According to PixLoc \cite{sarlin21pixloc}, the prescribed method for relocalization is hloc w/ SuperPoint + SuperGlue \cite{sarlin2019coarse, sarlin20superglue, detone18superpoint} followed by refinement step using feature-metric alignment. We use both these methods separately in PixTrack. As Pixloc was meant for visual localization, we observed a few limitations with out of the box usage of Pixloc for object-tracking. 

\subsection{In-plane rotations:}
We observe in our experiments that descriptors like SuperPoint \cite{detone18superpoint} and the features extracted from S2DNet \cite{Germain2020S2DNet} or UNet \cite{ronneberger2015u} are not robust to in-plane rotations. To combat this, we either need better descriptors or we need to augment the reference frames with in-plane rotations. This is undesirable as it'll increase the size of the SfM “features” and “matches” data-structures at $O(S*N)$ and $O(S^2*N^2)$ rates respectively where S is the number of in-plane rotations and N is the number of reference frames.

\begin{table*}[ht]
\label{ycb_results}
\centering
\begin{tabular}{lrcccccccc}
\hline
\\[-1em]
\multicolumn{1}{c}{\textbf{}} & \textbf{} & \multicolumn{4}{c}{\textbf{RGB}} & \multicolumn{4}{c}{\textbf{RGB + Depth}} \\ \hline
\\[-1em]
\multicolumn{2}{l}{Thresholds} & \multicolumn{2}{c}{15\%} & \multicolumn{2}{c}{20\%} & \multicolumn{2}{c}{15\%} & \multicolumn{2}{c}{20\%} \\ \hline
\\[-1em]
\multicolumn{2}{l}{\textbf{Object ID}} & \textbf{ADD} & \textit{\textbf{ADD-S}} & \textbf{ADD} & \textit{\textbf{ADD-S}} & \textbf{ADD} & \textit{\textbf{ADD-S}} & \textbf{ADD} & \textit{\textbf{ADD-S}} \\ \hline
\\[-1em]
\multicolumn{2}{l}{Cracker Box} & 0.97 & 0.99 & 0.99 & 0.99 & \textbf{0.99} & 0.99 & 0.99 & 0.99 \\
\\[-1em]
\multicolumn{2}{l}{Mustard Bottle} & 0.72 & 0.87 & 0.79 & 0.91 & 0.72 & \textbf{0.99} & \textbf{0.88} & \textbf{0.99} \\
\\[-1em]
\multicolumn{2}{l}{Sugar Box} & 0.75 & 0.98 & 0.86 & 0.99 & \textbf{0.89} & \textbf{0.99} & \textbf{0.97} & 0.99 \\
\\[-1em]
\multicolumn{2}{l}{Power Drill} & 0.89 & 0.99 & 0.96 & 0.99 & \underline{0.87} & 0.99 & \underline{0.93} & 0.99 \\
\\[-1em]
\multicolumn{2}{l}{Bleach Cleanser} & 0.63 & 0.79 & 0.67 & 0.88 & \textbf{0.73} & \textbf{0.91} & \textbf{0.78} & \textbf{0.94} \\ \hline
\\[-1em]
\multicolumn{2}{l}{\textbf{Average}} & 0.79 & 0.93 & 0.86 & 0.95 & \textbf{0.84} & \textbf{0.98} & \textbf{0.91} & \textbf{0.98} \\ \hline
\\[-1em]
\end{tabular}
\caption{YCB-Video results of objects on 3 videos each. We run on two input modes, RGB and RGBD while evaluating the accuracy. The ADD and ADD-S metrics are vertex distances compared to object-diameter percentages. We observe boosts in accuracy when we use depth as an additional feature. Regressions are highlighted for the power-drill, which is relatively small, as accuracy is high for both modes.}
\label{table:ycb_res}
\end{table*}

\subsection{Reference frame selection:}

Optimizing from the closest reference frame assumes that features at projected SfM points in both reference and query images are proximal. As the true pose of the query frame diverges from reference frames, feature-distances increases. Discontinuity in features arises while switching the closest reference frame, causing jitters in tracking during transitions.To ensure smooth tracking, we need a dense set of reference frames such that we are never outside the acceptable pose difference in reference and query frames. Choosing multiple reference frames makes the search slower, which isn't ideal to tracking. 

\subsection{SfM key-point visibility masks:}
When choosing the closest reference frame, visual localization methods make an assumption regarding the visibility of SfM key-points that the visibility mask for SfM key-points in the chosen reference frame and the true visibility mask in the query frame are mostly the same. Although, as the query frame moves further away from the reference frame, the number of False Positives and False Negatives for key-point visibility increases. A robust loss \cite{BarronCVPR2019} could help ignore the effects of the residuals from these False-Positives on visibility, the effectiveness of such a solution will decrease the further the query frame moves from the reference frame. 
\label{subsec:tracking}

\section{Proposed Method: PixTrack}
\label{subsec:object_nerf}
To overcome the limitations mentioned in \ref{subsec:tracking}, we propose using a Neural Radiance Field \cite{mildenhall2020nerf} as the canonical 3D representation of the object and re-purpose Pixloc\cite{sarlin21pixloc} as a feature-metric pose-optimizer. We use the pose from the previous query frame to synthesize a novel view, which we use as the reference frame for PixLoc. Inherently, this method renders reference frames on-the-fly that synchronously rotate in-plane with the query frame. This gets rid of the need to store any reference frames, which makes it memory efficient. Furthermore, these reference frames are very close to the true pose of the query frames, which makes the extracted reference and query features close at the projected points of interest. Since it is trivial to get the depth values for a NeRF view, we use the depth to provide the segmentation mask for RGB and help us generate geometrically accurate key-point visibility masks.


\subsection{Feature-metric alignment}
For each query image $I_q$, a corresponding reference image $I_k$ is generated by the object NeRF using the previous frame's pose. Following PixLoc \cite{sarlin21pixloc}, we extract a hierarchy of features using a CNN for $I_q$ and $I_k$. At each level $l \in \{L, ..., 1\}$ of the hierarchy, $C_l$ dimensional feature maps $F^{l} \in \mathbb{R}^{W_{l} \times H_{l} \times C_{l}}$ are extracted. 
The goal of the feature-metric optimization is to find a pose $(R, t)$ that minimizes the difference in appearance between the query and reference image in feature-space. When depth is available, we add an additional term that minimizes the distance between the query depth map $D_q$ and the visible 3D SfM points.
For a given feature level $l$ and each 3D point $i$ observed in the current reference frame, the feature-metric and depth residuals are defined as:

\begin{equation}
    r_{feat}^{i}=F_{q}^{l}[p_{q}^{i}]-F_{k}^{l}[p_{k}^{i}]\in \mathbb{R}^{C_{l}}
\end{equation}
\begin{equation}
    r_{depth}^{i}=D_{q}[p_{q}^{i}]-Z_{q}^{i}\in \mathbb{R} 
\end{equation}
Here, $p^{i}_{q}=\prod (RP^i+t)$ is the camera projection of $i$ onto the query frame, $\left[ \cdot  \right]$ denotes sub-pixel interpolation of at the projected point and $Z_q^i$ is the z-coordinate of $i$ in the query frame.

The total error over all points is given by 
\begin{equation}
    E_l(R, t)=\sum_{i}w^i\rho(\left\| r^i \right\|^2)
\end{equation}

Here, $\rho$ is a robust cost function with derivative $\rho'$ and $w^i$ is a per-residual weight. We predict occlusion regions in the query image by computing discontinuities with respect to the canonical object-NeRF's rendered depth. We mask out the residuals of points lying within the occlusion mask through $w^i$ such that the residuals are only applied to points projected to the visible regions of the object. The feature-metric and depth residuals are stacked into $r \in \mathbb{R}^{N(C+1)}$.  Each pose update $\delta \in \mathbb{R}^{6}$ is parameterized on the
$SE(3)$ manifold using its Lie algebra. The Jacobians for feature-metric and depth residuals are defined as:
\begin{equation}
    J_{feat}^{i}=\frac{\partial r_{feat}}{\partial \delta}=\frac{\partial F_q}{\partial p_{q}^{i}}\frac{\partial p_{q}^{i}}{\partial \delta}
\end{equation}
\begin{equation}
    J_{depth}^{i}=\frac{\partial r_{depth}}{\partial \delta}=\frac{\partial D_q}{\partial p_{q}^{i}}\frac{\partial p_{q}^{i}}{\partial \delta}-\frac{\partial Z_q^i}{\partial \delta}
\end{equation}
The jacobians are similarly stacked into $J \in \mathbb{R}^{N(C+1)}$. Given $W=diag_i(w^i\rho')$, the resulting Hessian matrix is:
\begin{equation}
    H=J^{\top}WJ
\end{equation}
We follow the optimization in \cite{sarlin21pixloc} to iteratively minimize the nonlinear least-squares cost using the Levenberg-Marquardt algorithm.

\subsection{Data collection protocol for Object-NeRF}
For a chosen object, we place the object in the center of a turntable with an aruco board on the rotating face. An iPhone 12 pro is mounted on a flexible tripod to collect images. With the focus and exposure fixed, RGB stills are captured at periodic intervals while the turntable is rotating. We repeat this process at three different mount elevations. We typically collect around 150 images of the object at a resolution of 4032x3024 pixels .

\subsection{SfM pipeline}
It is common practice to use COLMAP \cite{schoenberger2016sfm, schoenberger2016mvs} to generate reference camera poses and 3D maps for use in localization, as well as supporting NeRF creation. We observe that in order to capture high quality object-level NeRFs, and to perform robust localization using such models, it is imperative to have a highly robust and accurate SfM pipeline. Towards this, we start by employing an enhanced variant of COLMAP, hloc + SuperGlue \cite{sarlin2019coarse, detone18superpoint, sarlin20superglue} to create our SfM models. We further use PixSFM \cite{lindenberger2021pixsfm} to refine the SfM models using Feature-metric Key-point Adjustment and Feature-metric Bundle Adjustment, which merges incorrectly split tracks and reduces re-projection error.

\subsection{Extracting the Object NeRF}
After running SFM, we train a NeRF that contains both the Object and the turntable. 
To subtract out the turntable, we deploy a simple NeRF differencing algorithm inspired by GIRAFFE\cite{Niemeyer2020GIRAFFE} . We first train a NeRF on images only containing the turntable. 
Let's call the Object-NeRF with the turntable as $S_o$ and the turntable NeRF as $S_b$. 
Using instant-NGP \cite{mueller2022instant}, we first train the background NeRF $f_b$ using $S_b$. Next, we use NeRF compositing to train a “difference NeRF” $f_o$ using $S_o$. Consider the volume density $\sigma_b$ and the color $c_b$ for NeRF $f_b$. We learn $f_o$ with volume density $\sigma_o$ and color $c_o$  using the following composition:
\begin{equation}
    \sigma = \sigma_b + \sigma_o 
\end{equation}
\begin{equation}
c = \frac{\sigma_b * c_b + \sigma_o * c_o}{\sigma}
\end{equation}

While training $f_o$, we use composed volume density $\sigma$ and color $c$ to compute the final rgb pixel values. The weights for $f_b$ are frozen.
We filter out background 3D points from the SfM by fixing the camera poses and re-triangulating \cite{lindenberger2021pixsfm} using training views rendered by $f_o$.

\graphicspath{ {images/} }

\subsection{Efficient implementation}
\label{subsec:efficient_implementation}
There are several bottle necks that slow down tracking. To avoid rendering novel views every frame, we cache the reference frame features and re-use them for any pose within thresholds for euclidean distance for translation and geodesic distance for rotation. We also constrain the pose search by generating an approximate segmentation mask for the query frame using the object-NeRF's previous pose, and manage to eliminate most of the background. This adds to robustness, and also helps with early stopping of the solver.


\section{Experimental Results}
\label{sec:experiments}

To assess the performance of Pixtrack, we evaluated it on the YCB-Video dataset using the ADD and ADD-S metrics. The high-resolution CAD models provided by the YCB-object dataset \cite{YCB-object} were employed in conjunction with virtual camera-captured images to train a NeRF for each object.

The ADD and ADD-S metrics were utilized to determine pose-tracking accuracy, with threshold values being the percentages of object diameters against vertex l2 distance as proposed in HybridPose\cite{song2020hybridpose}. 
The pipeline relies on ground-truth poses for initialization and relocalization recovery. 

The RGB-only pipeline demonstrates satisfactory performance in scenarios with minimal occlusions and background objects. Remarkably, the tracker's pixel-level accuracy can be observed in our video-outputs.
As evidenced in Table \ref{table:ycb_res}, incorporating depth as a separate feature distinctly enhances performance, with the exception of the Power-Drill, which experiences an inconsequential regression. Our findings suggest that depth utilization for optimization and occlusion-detection improves performance for objects that are not optimally suited for monocular RGB alone. 
We also did tests with just using depth, which worked satisfactorily, although we did observe jitter and shape-symmetry issues. 
Without the Jacobian terms in equation-5, and using depth just like another feature, we don't observe the performance boost, which proves the analytical solution is indeed correct.  

\section{Conclusion and Future Work}
\label{sec:future}
We describe an end-to-end 6-DOF object pose tracking system for monocular RGB and RGB-D that uses novel view synthesis with deep feature-metric alignment to achieve jitter-free tracking with no filtering. We address key limitations sfm re-localization based tracking: we overcome poor matching with in-plane rotations, sparsely distributed reference frames and poor 3d key-point selection by using a NeRF as the canonical representation of the object. The object-NeRF enables us to gracefully handle in-plane rotations since it is geometrically consistent; we can produce photo-realistic reference frames from anywhere in pose space; and using the density field, we can accurately filter out 3d points for accurate feature-metric alignment. Further, we extend the feature-metric optimization and add a depth loss term that further improves performance, not relying on an extra ICP step. Like \cite{wen2023bundlesdf} we'd like to use a segmentation-network to segment the query-frames instead of using the nerf-depth as a mask.

Future work could include additional regularization terms on object velocity, acceleration, to mitigate the number of re-localizations while tracking. Training a feature detector derived from the features in the Multi-resolution hash-map of INGP is also an interesting direction. 


{\small
\bibliographystyle{ieee_fullname}
\bibliography{egbib}

\begin{thebibliography}{10}\itemsep=-1pt

\bibitem{BarronCVPR2019}
Jonathan~T. Barron.
\newblock A general and adaptive robust loss function.
\newblock {\em CVPR}, 2019.

\bibitem{YCB-object}
Berk Calli, Arjun Singh, Aaron Walsman, Siddhartha Srinivasa, Pieter Abbeel,
  and Aaron~M. Dollar.
\newblock The ycb object and model set: Towards common benchmarks for
  manipulation research.
\newblock In {\em 2015 International Conference on Advanced Robotics (ICAR)},
  pages 510--517, 2015.

\bibitem{deeplabv3plus2018}
Liang-Chieh Chen, Yukun Zhu, George Papandreou, Florian Schroff, and Hartwig
  Adam.
\newblock Encoder-decoder with atrous separable convolution for semantic image
  segmentation.
\newblock In {\em ECCV}, 2018.

\bibitem{detone18superpoint}
Daniel DeTone, Tomasz Malisiewicz, and Andrew Rabinovich.
\newblock Superpoint: Self-supervised interest point detection and description.
\newblock In {\em CVPR Deep Learning for Visual SLAM Workshop}, 2018.

\bibitem{Germain2020S2DNet}
Hugo Germain, Guillaume Bourmaud, and Vincent Lepetit.
\newblock S2dnet: Learning image features for accurate sparse-to-dense
  matching.
\newblock In {\em European Conference on Computer Vision (ECCV)}, 2020.

\bibitem{kaiminghe_MaskRCNN}
Kaiming He, Georgia Gkioxari, Piotr Dollár, and Ross Girshick.
\newblock Mask r-cnn, 2017.

\bibitem{NeRFPose}
Fu Li, Hao Yu, Ivan Shugurov, Benjamin Busam, Shaowu Yang, and Slobodan Ilic.
\newblock Nerf-pose: A first-reconstruct-then-regress approach for
  weakly-supervised 6d object pose estimation, 2022.

\bibitem{lindenberger2021pixsfm}
Philipp Lindenberger, Paul-Edouard Sarlin, Viktor Larsson, and Marc Pollefeys.
\newblock {Pixel-Perfect Structure-from-Motion with Featuremetric Refinement}.
\newblock In {\em ICCV}, 2021.

\bibitem{mildenhall2020nerf}
Ben Mildenhall, Pratul~P. Srinivasan, Matthew Tancik, Jonathan~T. Barron, Ravi
  Ramamoorthi, and Ren Ng.
\newblock Nerf: Representing scenes as neural radiance fields for view
  synthesis.
\newblock In {\em ECCV}, 2020.

\bibitem{mueller2022instant}
Thomas M\"uller, Alex Evans, Christoph Schied, and Alexander Keller.
\newblock Instant neural graphics primitives with a multiresolution hash
  encoding.
\newblock {\em ACM Trans. Graph.}, 41(4):102:1--102:15, July 2022.

\bibitem{Niemeyer2020GIRAFFE}
Michael Niemeyer and Andreas Geiger.
\newblock Giraffe: Representing scenes as compositional generative neural
  feature fields.
\newblock In {\em Proc. IEEE Conf. on Computer Vision and Pattern Recognition
  (CVPR)}, 2021.

\bibitem{ronneberger2015u}
Olaf Ronneberger, Philipp Fischer, and Thomas Brox.
\newblock U-net: Convolutional networks for biomedical image segmentation.
\newblock In {\em International Conference on Medical image computing and
  computer-assisted intervention}, pages 234--241. Springer, 2015.

\bibitem{sarlin2019coarse}
Paul-Edouard Sarlin, Cesar Cadena, Roland Siegwart, and Marcin Dymczyk.
\newblock From coarse to fine: Robust hierarchical localization at large scale.
\newblock In {\em CVPR}, 2019.

\bibitem{sarlin20superglue}
Paul-Edouard Sarlin, Daniel DeTone, Tomasz Malisiewicz, and Andrew Rabinovich.
\newblock {SuperGlue}: Learning feature matching with graph neural networks.
\newblock In {\em CVPR}, 2020.

\bibitem{sarlin21pixloc}
Paul-Edouard Sarlin, Ajaykumar Unagar, Måns Larsson, Hugo Germain, Carl Toft,
  Victor Larsson, Marc Pollefeys, Vincent Lepetit, Lars Hammarstrand, Fredrik
  Kahl, and Torsten Sattler.
\newblock {Back to the Feature: Learning Robust Camera Localization from Pixels
  to Pose}.
\newblock In {\em CVPR}, 2021.

\bibitem{schoenberger2016sfm}
Johannes~Lutz Sch\"{o}nberger and Jan-Michael Frahm.
\newblock Structure-from-motion revisited.
\newblock In {\em Conference on Computer Vision and Pattern Recognition
  (CVPR)}, 2016.

\bibitem{schoenberger2016mvs}
Johannes~Lutz Sch\"{o}nberger, Enliang Zheng, Marc Pollefeys, and Jan-Michael
  Frahm.
\newblock Pixelwise view selection for unstructured multi-view stereo.
\newblock In {\em European Conference on Computer Vision (ECCV)}, 2016.

\bibitem{song2020hybridpose}
Chen Song, Jiaru Song, and Qixing Huang.
\newblock Hybridpose: 6d object pose estimation under hybrid representations,
  2020.

\bibitem{OnePose}
Jiaming Sun, Zihao Wang, Siyu Zhang, Xingyi He, Hongcheng Zhao, Guofeng Zhang,
  and Xiaowei Zhou.
\newblock Onepose: One-shot object pose estimation without cad models, 2022.

\bibitem{wen2023bundlesdf}
Bowen Wen, Jonathan Tremblay, Valts Blukis, Stephen Tyree, Thomas Muller, Alex
  Evans, Dieter Fox, Jan Kautz, and Stan Birchfield.
\newblock Bundlesdf: Neural 6-dof tracking and 3d reconstruction of unknown
  objects.
\newblock {\em CVPR}, 2023.

\bibitem{RNNPose}
Yan Xu, Kwan-Yee Lin, Guofeng Zhang, Xiaogang Wang, and Hongsheng Li.
\newblock Rnnpose: Recurrent 6-dof object pose refinement with robust
  correspondence field estimation and pose optimization, 2022.

\bibitem{iNeRF}
Lin Yen-Chen, Pete Florence, Jonathan~T. Barron, Alberto Rodriguez, Phillip
  Isola, and Tsung-Yi Lin.
\newblock Inerf: Inverting neural radiance fields for pose estimation, 2020.

\end{thebibliography}
}


\end{document}